\documentclass[conference]{IEEEtran}

\usepackage[utf8]{inputenc}
\usepackage[T1]{fontenc}
\usepackage{cite}
\usepackage{amsmath,amssymb,amsfonts}
\usepackage{mathtools}
\usepackage{amsthm}
\usepackage{graphicx}
\usepackage{textcomp}
\usepackage{xcolor}
\usepackage[font=small]{caption}
\usepackage{subcaption}
\usepackage{tikz,pgfplots}
\usepackage{pgfplotstable} 
\pgfplotsset{compat=1.18}
\usepackage{booktabs}
\usepackage{multirow}
\usepackage{array}
\usepackage{url}
\usepackage{hyperref}
\usepackage{nicefrac}
\usepackage{adjustbox}
\usepackage{xspace}
\usepackage{wrapfig}
\usepackage{algorithm}
\usepackage{algpseudocode}
\usepackage{enumitem}
\usepackage[capitalize,noabbrev]{cleveref}
\usepackage{tcolorbox}
\tcbuselibrary{listings,breakable}
\usepackage{verbatim}
\usepackage[table]{xcolor}

\let\citep\cite

\newcommand{\citet}[1]{\cite{#1}}

\newcommand{\framework}{\textsc{ScenarioDiff}\xspace}

\definecolor{rankfirst}{RGB}{192,0,0}      
\definecolor{ranksecond}{RGB}{0,90,180}    %
\newcommand{\first}[1]{\textcolor{rankfirst}{\textbf{#1}}}
\newcommand{\second}[1]{\textcolor{ranksecond}{\textbf{#1}}}
\newcommand{\third}[1]{\textbf{#1}}

\newcommand{\firstlegend}[1]{\protect\textcolor{rankfirst}{#1}}
\newcommand{\secondlegend}[1]{\protect\textcolor{ranksecond}{#1}}
\newcommand{\thirdlegend}[1]{\protect\textbf{#1}}

\newtcolorbox{promptbox}[1][]{
  breakable,
  colback=gray!3,
  colframe=black!50,
  boxrule=0.5pt,
  arc=2pt,
  left=6pt,
  right=6pt,
  top=6pt,
  bottom=6pt,
  fonttitle=\bfseries,
  title=#1
}

\hypersetup{
  hidelinks,
  colorlinks=false,
  pdfborder={0 0 0}
}

\theoremstyle{plain}
\newtheorem{theorem}{Theorem}[section]

\theoremstyle{definition}

\theoremstyle{remark}
\newtheorem{remark}[theorem]{Remark}

\newcommand{\todo}[2][]{ }

\title{\framework: A Scenario-level Guidance Framework for Multimodal Time Series Forecasting}

\author{
    \IEEEauthorblockN{
        Tuan-Binh Tran$^{1}$, 
        Dat Nguyen-Cong$^{2}$, 
        Duc-Trong Le$^{3}$,
        Thanh Trung Huynh$^{1}$,
        Tung Kieu$^{4}$,
    } 
    \IEEEauthorblockA{
        $^1$\textit{VinUniversity, Hanoi, Vietnam}, $^2$\textit{FPT Software AI Center, FPT Corporation, Hanoi, Vietnam}, \\$^3$\textit{VNU University of Engineering and Technology, Hanoi, Vietnam},
        $^4$\textit{Aalborg University, Aalborg, Denmark} \\
        $^1$\{binh.tt2, trung.ht\}@vinuni.edu.vn, 
        $^2$dat27072002@gmail.com,
        $^3$trongld@vnu.edu.vn
        $^4$tungkvt@cs.aau.dk
    }
} 

\begin{document}

\maketitle

\begin{abstract}
    Textual context such as news, reports, and logs can provide valuable signals for time series forecasting, especially when future dynamics are driven by external events that are not yet visible in historical values. Existing multimodal forecasting methods often either ask large language models (LLMs) to predict numerical values directly or fuse text and time series implicitly, making contextual influence difficult to interpret and control. We propose \framework, a hierarchical contextual reasoning framework for multimodal time series forecasting under noisy and weakly aligned documents. \framework organizes contextual information into three levels: a \emph{Historical Context Agent} extracts stepwise evidence from raw documents, a \emph{Scenario Agent} produces a qualitative scenario description for the forecast horizon, and an \emph{Anchor Guidance Agent} generates sparse anchor points for event-relevant future regions. These structured signals condition a \emph{Multimodal Diffusion Transformer}, while \emph{Anchor Blended Sampling} locally refines generated trajectories without retraining. Experiments on the \textbf{Time-MMD} benchmark show that \framework is especially effective in event-driven domains, demonstrating the value of explicit hierarchical scenario guidance for multimodal time series forecasting. Our full implementation is available at \url{https://anonymous.4open.science/r/ScenarioDiff_ICDM-2C4C}.
\end{abstract}

\begin{IEEEkeywords}
time series forecasting, multimodal forecasting, diffusion models, large language models, scenario guidance
\end{IEEEkeywords}
\section{Introduction}
\label{introduction}

\begin{figure}
    \centering
    \includegraphics[width=1\linewidth]{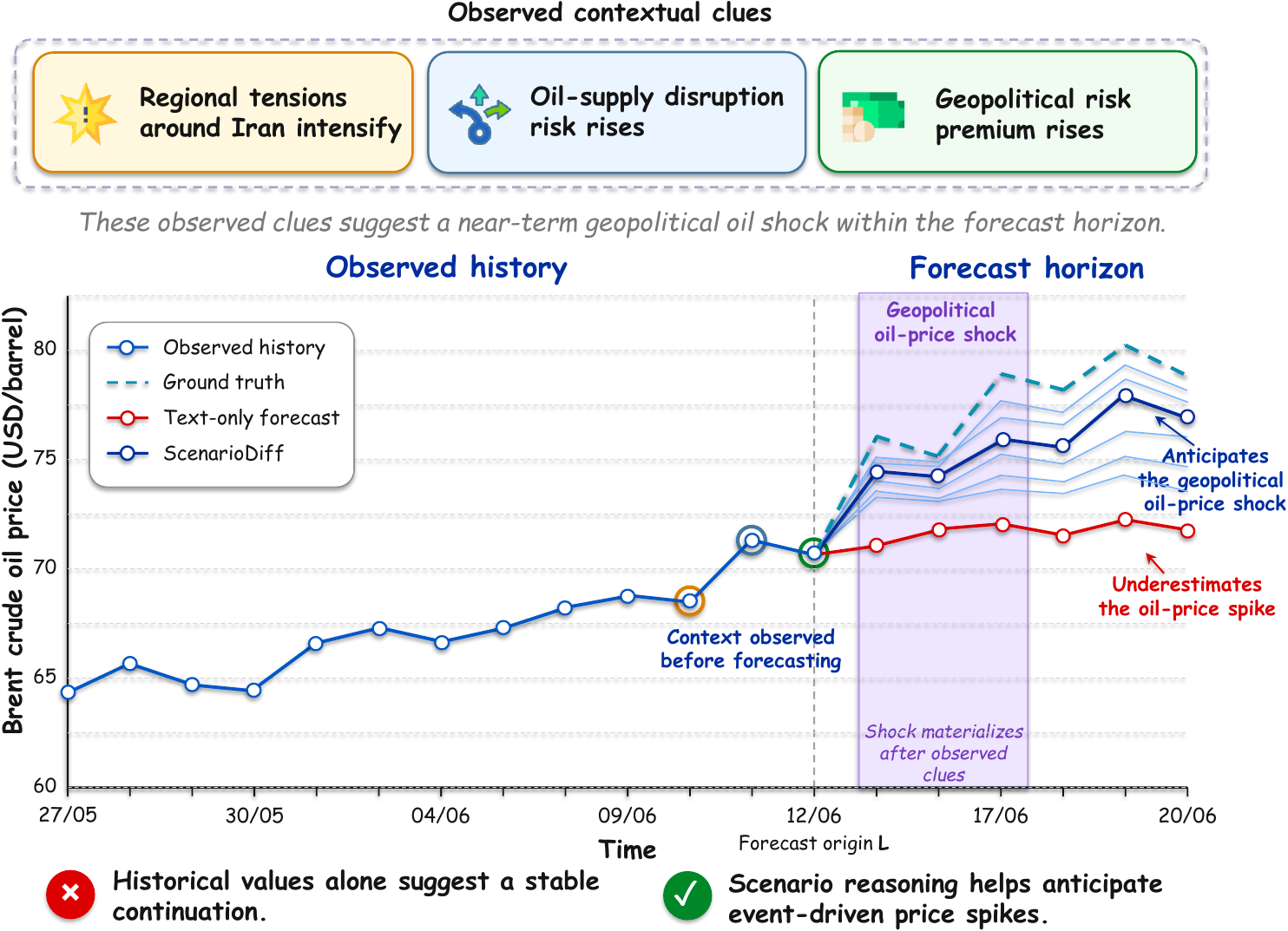}
    \caption{
        \textbf{Motivating example.} Brent crude oil prices around the Israel--Iran conflict in June 2025 illustrate the need for scenario-level reasoning. Before the forecast origin $L$, historical prices show only a moderate upward trend, while available contextual clues indicate rising regional tensions, oil-supply disruption risk, and geopolitical risk premium. These signals suggest a near-term oil-price shock that is not fully reflected in the numerical history. \framework uses the observed context to form a disruption scenario and produces trajectories that better follow the realized price increase.    
    }   
    \label{fig:motivation}
    \vspace{-10pt}
\end{figure}

Time series forecasting (TSF) supports decision-making in domains such as energy~\citep{DBLP:conf/ijcai/TruongMTCR13}, economics~\citep{DBLP:conf/kdd/ZhangHWLQCX0WW25}, healthcare~\citep{DBLP:conf/aaai/NoroozizadehKW26}, and transportation~\citep{DBLP:journals/pvldb/KieuKHYJL24}. Despite its importance, most forecasting models operate primarily on numerical histories. In practice, however, human experts rarely interpret temporal signals in isolation. An energy engineer may analyze demand traces together with weather forecasts and maintenance logs, while a financial analyst may read price series alongside earnings reports and macroeconomic news. Such contextual information can explain historical dynamics and, more importantly, indicate future developments that are not yet visible from numerical values alone. A motivating example is shown in Figure~\ref{fig:motivation}, where contextual evidence provides useful clues about future dynamics beyond what is visible from the numerical history alone.

This observation has motivated growing interest in \emph{multimodal} time series forecasting (MTSF), where numerical histories are paired with contextual text such as news, reports, and logs. Existing studies show that textual information can improve forecasting across domains~\citep{DBLP:conf/nips/LiuXZKKSSCW0P24,DBLP:journals/corr/abs-2502-08942,DBLP:conf/aaai/Wang0W0ZWZL25,DBLP:journals/corr/abs-2505-10774}. However, real-world text is often noisy, redundant, weakly aligned with timestamps, and only partially relevant to the target series. Under such conditions, implicit text--time fusion makes it difficult to identify which evidence influences the forecast and how strongly the model should trust it.

Large language models (LLMs) are a natural tool for processing unstructured context because they can summarize, filter, and reason over textual evidence. Recent studies have explored two main directions. The first direction prompts or reprograms numerical histories into language-like inputs and asks LLMs to predict future values directly~\citep{DBLP:conf/iclr/0005WMCZSCLLPW24}. While appealing, this places long-horizon numerical generation on models primarily optimized for language understanding rather than precise temporal extrapolation~\citep{DBLP:conf/kdd/ZhengDZY0M025}. The second direction incorporates text through multimodal fusion, cross-attention, or agent-based summarization pipelines~\citep{DBLP:conf/nips/LiuXZKKSSCW0P24,DBLP:journals/corr/abs-2502-08942,DBLP:conf/aaai/Wang0W0ZWZL25,DBLP:journals/corr/abs-2505-10774}. These methods are effective, but often assume clean text--timestamp alignment or absorb contextual evidence implicitly in latent space.

An additional challenge is that contextual text may describe not only smooth trend continuation, but also event-driven deviations such as policy interventions, outages, or abrupt demand shifts. Capturing such cases requires future-oriented hypotheses that can guide a numerical forecaster without asking an LLM to directly output numbers. It also requires a mechanism for translating coarse textual expectations into localized constraints when abrupt changes are likely.

In this work, we propose \framework, a hierarchical contextual reasoning framework for MTSF. The key idea is to organize contextual information into three complementary levels of guidance, each serving a different role in guiding the forecast. First, a frozen \emph{Historical Context Agent} compresses raw documents into \emph{stepwise context summaries} aligned with the observed history, grounding the model in document-level evidence. Second, a frozen \emph{Scenario Agent} synthesizes the historical series and these summaries into a \emph{scenario description}, which serves as a qualitative prior over the forecast horizon. Third, an \emph{Anchor Guidance Agent} converts the available context into sparse \emph{anchor points}, providing time-localized guidance for future regions where abrupt changes are likely. This hierarchy separates past evidence, future hypotheses, and local trajectory constraints, making contextual influence more explicit and modular than implicit fusion.

We instantiate the forecaster as a \emph{Multimodal Diffusion Transformer}, which models future trajectories through iterative denoising conditioned on the structured signals produced by the agents. The stepwise context summaries and scenario description guide the base diffusion process, allowing the model to combine numerical history with scenario-level textual evidence without asking LLMs to directly output numerical forecasts. To make anchor points actionable, we introduce \emph{Anchor Blended Sampling}, an inference-time refinement procedure that locally edits anchor-relevant regions using a distance-to-band objective while preserving non-anchor regions through blended diffusion. Thus, anchor points act as soft local constraints rather than hard overrides, encouraging event-consistent trajectories without retraining the base forecaster.

We evaluate \framework on the \textbf{Time-MMD} benchmark~\citep{DBLP:conf/nips/LiuXZKKSSCW0P24}. Results show that \framework achieves strong performance, with the clearest gains in event-driven domains where textual evidence contains actionable signals about future dynamics. This supports our central hypothesis that explicit hierarchical scenario guidance is most beneficial when future changes are driven by external events weakly reflected in the numerical history.

In summary, our main contributions are as follows:
    (i) We introduce \framework, a hierarchical contextual reasoning framework that transforms noisy textual documents into three levels of guidance for MTSF: historical evidence, future scenario, and local anchor constraints.
    (ii) We integrate these structured signals into a Multimodal Diffusion Transformer, enabling probabilistic forecasting conditioned on explicit historical and scenario-level context.
    (iii) We propose Anchor Blended Sampling, which connects anchor points from the Anchor Guidance Agent with local post-hoc trajectory refinement without retraining the diffusion forecaster.
    (iv) We provide empirical analysis on \textbf{Time-MMD}, showing that the main gains concentrate in event-driven domains where textual evidence carries actionable future signals.

\section{Related Work}
\label{related_work}

\textbf{LLM-Based Time Series Forecasting.}
Recent work has explored large language models (LLMs) as sequence priors for time series analysis and forecasting. \texttt{OFA}~\cite{DBLP:conf/nips/ZhouNW0023} studies how pretrained language-model backbones can be transferred to time series tasks. \texttt{Time-LLM}~\cite{DBLP:conf/iclr/0005WMCZSCLLPW24} reprograms numerical histories into language-like representations, enabling frozen LLMs to perform few-shot and zero-shot forecasting. \texttt{CALF}~\cite{DBLP:conf/aaai/LiuG0LBR0X25} further improves this paradigm by aligning temporal tokens with textual representations. These methods demonstrate the potential of LLMs to provide semantic priors for forecasting. However, they often require LLMs to support numerical extrapolation, which is not ideal for precise long-horizon prediction due to the mismatch between language modeling and fine-grained temporal generation. In contrast, \framework uses LLMs for producing contextual information rather than direct numerical generation: the agents summarize historical evidence, produce scenario descriptions, and extract anchor points, while a diffusion forecaster handles probabilistic generation.

\textbf{Multimodal Time Series Forecasting.}
Multimodal time series forecasting incorporates contextual text such as news, reports, logs, and event descriptions into numerical forecasting models. \texttt{Time-MMD}~\cite{DBLP:conf/nips/LiuXZKKSSCW0P24} introduces a multi-domain benchmark and shows that paired textual information can improve forecasting across domains. \texttt{TaTS}~\cite{DBLP:journals/corr/abs-2502-08942} aligns text with each timestep and uses text embeddings as additional temporal features. \texttt{TimeCAP}~\cite{DBLP:conf/aaai/LeeYS0C25} adopts an agent-based pipeline to summarize and contextualize time series before prediction, while \texttt{ChatTime}~\cite{DBLP:conf/aaai/Wang0W0ZWZL25} jointly models numerical and textual inputs in a unified \texttt{Transformer}. Although effective, these approaches often assume clean text--timestamp alignment or fuse text implicitly in latent space, making it difficult to inspect which evidence affects the forecast under noisy, weakly aligned, multi-document inputs. \framework differs by converting raw documents into explicit stepwise context summaries, a scenario description, and anchor points, making the contextual roles more modular and interpretable.

\textbf{Diffusion Models for Time Series Forecasting.}
Diffusion models have become a strong class of probabilistic forecasters, generating future trajectories through iterative denoising. Existing methods include one-shot horizon generation~\cite{DBLP:conf/nips/TashiroSSE21,DBLP:conf/icml/ShenK23} and autoregressive diffusion-based prediction~\cite{DBLP:conf/icml/RasulSSV21}. Subsequent work improves long-horizon forecasting with multi-resolution denoising, structured backbones, and retrieval-augmented conditioning~\cite{DBLP:conf/iclr/ShenCK24,DBLP:journals/tmlr/AlcarazS23,DBLP:conf/nips/LiuY0H24}. Recent multimodal diffusion methods further inject timestamps and textual context through fusion, cross-attention, or classifier-free guidance~\cite{DBLP:journals/corr/abs-2504-19669,DBLP:journals/corr/abs-2512-07184}, while flow-matching models aim to reduce sampling cost~\cite{DBLP:conf/icml/LiuQSCY00L25,DBLP:conf/iclr/WuJQ0Y0G26}. Most prior methods for condition generation are mainly through architecture-level fusion and do not explicitly control how external context shapes the sampling. Our work builds on diffusion forecasting but differs in two aspects. First, the conditioning signals are not raw text or generic latent embeddings; instead, they are structured scenario-level signals that separately encode historical evidence and future hypotheses. Second, beyond architectural conditioning, we introduce an inference-time guidance mechanism that steers generated trajectories toward sparse anchor bands implied by contextual evidence. This provides an interpretable and flexible way to shape the forecasting distribution while remaining compatible with standard diffusion backbones.
\section{Preliminaries}
\subsection{Denoising Diffusion Probabilistic Models}
Denoising Diffusion Probabilistic Models (DDPMs)~\cite{DBLP:conf/nips/HoJA20} are generative models that iteratively add Gaussian noise to a clean sample, $\mathbf{x}^{(0)}$, over $T$ steps, following a Markov chain. The noise schedule, $\{\beta_t\}_{t=1}^T$, consists of $\beta_t \in (0,1)$, and the cumulative noise is represented by $\bar{\alpha}_t = \prod_{i=1}^t(1 - \beta_i)$. The forward diffusion process is described by:
\begin{align}
q(\mathbf{x}^{(1:T)}\mid\mathbf{x}^{(0)})
&= \prod_{t=1}^{T} q(\mathbf{x}^{(t)}\mid\mathbf{x}^{(t-1)}), \\
q(\mathbf{x}^{(t)}\mid\mathbf{x}^{(t-1)})
&= \mathcal{N}\big(\sqrt{1-\beta_t}\,\mathbf{x}^{(t-1)},\beta_t\mathbf{I}\big).
\end{align}
This process allows us to directly sample $\mathbf{x}^{(t)}$ from the clean sample $\mathbf{x}^{(0)}$ as:
\begin{equation}
    \mathbf{x}^{(t)} = \sqrt{\bar\alpha_t}\,\mathbf{x}^{(0)} + \sqrt{1-\bar\alpha_t}\,\boldsymbol{\epsilon}, \quad\quad \boldsymbol{\epsilon}\sim\mathcal{N}(0,\mathbf{I}).
\end{equation}
To reverse this diffusion chain, DDPM learns to progressively remove noise in each step. The reverse process is also modeled as a Markov chain and is defined as:
\begin{align}
p_\theta(\mathbf{x}^{(0:T)})
&= p_\theta(\mathbf{x}^{(T)})
\prod_{t=1}^{T} p_\theta(\mathbf{x}^{(t-1)}\mid\mathbf{x}^{(t)}), \\
p_\theta(\mathbf{x}^{(t-1)}\mid\mathbf{x}^{(t)})
&= \mathcal{N}\big(\mu(\mathbf{x}^{(t)},\mathbf{x}_\theta(\mathbf{x}^{(t)},t)),\Sigma_t\big).
\end{align}
Here, $p_\theta(\mathbf{x}^{(T)})$ is chosen as a standard Gaussian $\mathcal{N}(0,\mathbf{I})$, while $\mathbf{x}_\theta(\mathbf{x}^{(t)},t)$ is the predicted clean sample $\mathbf{x}^{(0)}$ from the noisy input $\mathbf{x}^{(t)}$, and $\Sigma_t$ is typically a fixed variance function.

\subsection{Conditional DDPM For Time Series Forecasting}
In time series forecasting, we extend the DDPM framework to handle conditional inputs. Let $\mathbf{c}$ represent the conditioning signals; the goal of the denoising network is to predict the future horizon, $\mathbf{x}^{(0)} \equiv \mathbf{x}_{L+1:L+H} = \langle \mathbf{x}_{L+1}, \mathbf{x}_{L+2}, \ldots \mathbf{x}_{L+H} \rangle$, based on the noisy future time series $\mathbf{x}^{(t)}$ and the condition $\mathbf{c}$. 
The denoising objective is formulated as:
\begin{equation}
    \mathcal{L}(\theta)
= \mathbb{E}_{\mathbf{x}^{(0)},t,\boldsymbol{\epsilon}}
\Big[
\big\|
\mathbf{x}^{(0)} -
\mathbf{x}_\theta(\mathbf{x}^{(t)}, t, \mathbf{c})
\big\|_2^2
\Big].
\end{equation}

Normally, $\mathbf{c}$ could be the lookback window $\mathbf{x}_{1:L}$.
However, in our setting, we also integrate textual context to enrich the condition.
Specifically, for each historical timestep $n\in\{1,\dots,L\}$, there is a local bag $\mathcal{M}_n=\{s_i\}_{i=1}^{|\mathcal{M}_n|}$, which contains textual descriptions of events occurring at that point.
\section{Methodology} 
\label{methodology}

\subsection{Framework Overview}
\begin{figure*}[ht]
    \centering
    \includegraphics[width=0.9\linewidth]{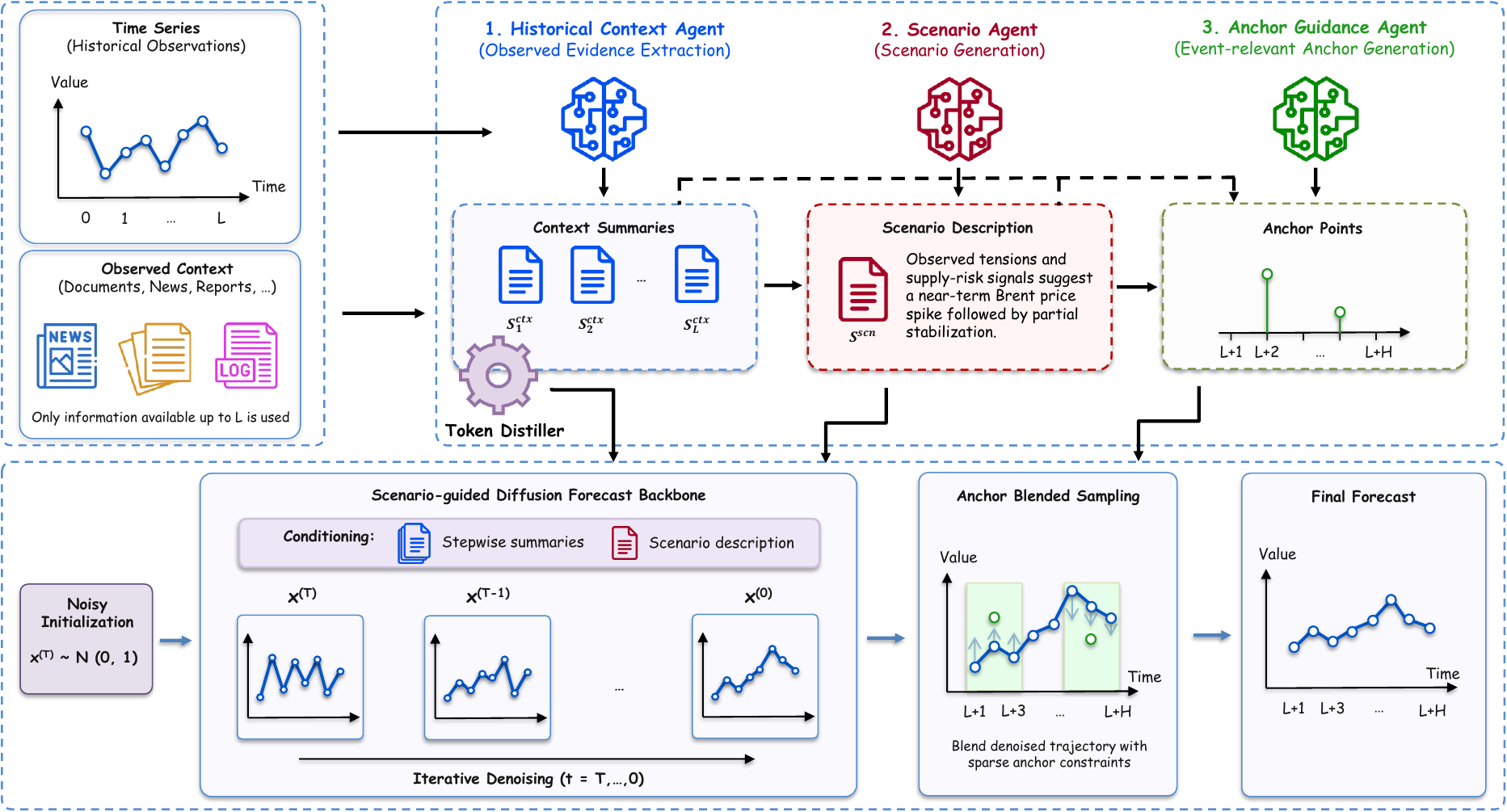}
    \caption{\framework overview.}
    \label{fig:full_pipeline}
    \vspace{-15pt}
\end{figure*}

Fig.~\ref{fig:full_pipeline} summarizes \framework. Let $\mathbf{x}_{1:L}\in\mathbb{R}^{L}$ denote the observed lookback window and $\mathbf{x}_{L+1:L+H}\in\mathbb{R}^{H}$ denote the future horizon. For each historical timestep $n\in\{1,\dots,L\}$, we assume access to an aligned document set $\mathcal{M}_n$. \framework organizes contextual reasoning into three hierarchical levels. First, a frozen \emph{Historical Context Agent} maps $\mathcal{M}_{1:L}$ to \emph{stepwise context summaries} $\mathbf{s}_{1:L}^{\mathrm{ctx}}$, grounding the model in document-aligned historical evidence. Second, a frozen \emph{Scenario Agent} maps $(\mathbf{x}_{1:L}, \mathbf{s}_{1:L}^{\mathrm{ctx}})$ to a \emph{scenario description} $\mathbf{s}^{\mathrm{scn}}$, projecting a coarse qualitative prior over the forecast horizon. Third, an \emph{Anchor Guidance Agent} maps $(\mathbf{x}_{1:L}, \mathbf{s}_{1:L}^{\mathrm{ctx}}, \mathbf{s}^{\mathrm{scn}})$ to sparse \emph{anchor points} $\mathcal{A}$, which provide localized constraints for inference-time trajectory refinement.

We instantiate the numerical forecaster as a conditional denoising diffusion model that estimates
\(
p\!\left(
\mathbf{x}_{L+1:L+H}
\mid
\mathbf{x}_{1:L}, \mathbf{s}_{1:L}^{\mathrm{ctx}}, \mathbf{s}^{\mathrm{scn}}
\right).
\)
Let $\mathbf{x}^{(0)}\equiv \mathbf{x}_{L+1:L+H}$ be the clean future target. The forward process corrupts it with Gaussian noise:
\(
q(\mathbf{x}^{(t)}\mid \mathbf{x}^{(0)})
=
\mathcal{N}\!\left(
\sqrt{\bar{\alpha}_t}\,\mathbf{x}^{(0)},
(1-\bar{\alpha}_t)\mathbf{I}
\right),
\)
or equivalently,
\(
\mathbf{x}^{(t)}
=
\sqrt{\bar{\alpha}_t}\,\mathbf{x}^{(0)}
+
\sqrt{1-\bar{\alpha}_t}\,\boldsymbol{\epsilon},
\qquad
\boldsymbol{\epsilon}\sim\mathcal{N}(\mathbf{0},\mathbf{I}).
\)
During denoising, the model receives the observed history $\mathbf{x}_{1:L}$, the noisy future $\mathbf{x}^{(t)}$, and the contextual conditions produced by the Historical Context Agent and the Scenario Agent. Starting from Gaussian noise, the reverse process generates a trajectory consistent with both numerical history and scenario-level context. Anchor points $\mathcal{A}$ are then used by Anchor Blended Sampling to locally refine the generated trajectory.

\subsection{Hierarchical Contextual Reasoning}


\paragraph{Historical Context Agent}

Raw documents aligned with each historical timestep are often verbose, redundant, and only weakly related to the forecasting target. Directly encoding such documents may obscure useful event signals. We therefore use a frozen \emph{Historical Context Agent} to compress each local document set into a concise summary.

For each timestep $n$, the agent takes $\mathcal{M}_n$ as input and produces a short summary $s_n^{\mathrm{ctx}}$. If no useful evidence is available, the agent emits a special \texttt{[NO\_EVENT]} token. Collecting these outputs yields the sequence
\(
\mathbf{s}_{1:L}^{\mathrm{ctx}} = \langle s_1^{\mathrm{ctx}}, \dots, s_L^{\mathrm{ctx}} \rangle.
\)
These summaries provide a temporally grounded semantic view of the observed history and serve as the main textual representation of past evidence.

To obtain continuous conditioning features, each summary is encoded by a text encoder $\mathrm{Enc}_{\mathrm{text}}(\cdot)$ followed by an average pooling as:
\begin{equation}
\mathbf{c}^{\mathrm{hist}}_n
=
\mathrm{AvgPool}\!\left(\mathrm{Enc}_{\mathrm{text}}(\mathbf{s}_n^{\mathrm{ctx}})\right),
\qquad
\mathbf{c}^{\mathrm{hist}}_n \in \mathbb{R}^{d_{\mathrm{emb}}}.
\end{equation}
We denote the encoded historical context sequence as $\mathbf{C}^{\mathrm{hist}}_{1:L} = \langle \mathbf{c}^{\mathrm{hist}}_1,\dots,\mathbf{c}^{\mathrm{hist}}_L\rangle$. The resulting sequence $\mathbf{C}^{\mathrm{hist}}_{1:L}$ is later injected into the historical tokens of the diffusion forecaster.

\paragraph{Scenario Agent}
While the stepwise context summaries capture fine-grained historical evidence, forecasting also benefits from a global qualitative hypothesis about the future horizon. We introduce a frozen \emph{Scenario Agent}, which generates a \emph{scenario description} from the observed history and the stepwise context summaries:
\(
\mathbf{s}^{\mathrm{scn}} = \mathrm{LLM}(\mathbf{x}_{1:L}, \mathbf{s}_{1:L}^{\mathrm{ctx}}).
\)
Because the same history-only procedure is used in both training and inference, this design avoids temporal leakage.

The scenario description is a short natural-language statement about expected future behavior, such as overall direction, volatility, or likely disruptions. We encode it with the same text encoder $\mathrm{Enc}_{\mathrm{text}}(\cdot)$:
\begin{align}
\mathbf{C}^{\mathrm{scn}} = \mathrm{Enc}_{\mathrm{text}}(\mathbf{s}^{\mathrm{scn}}),
\qquad
\mathbf{C}^{\mathrm{scn}} \in \mathbb{R}^{N_f \times d_{\mathrm{emb}}}.
\end{align}
This representation serves as a horizon-level prior for the diffusion forecaster.

\paragraph{Scenario Consistency Score}
For analysis only, we define a Scenario Consistency Score (SCS) to measure whether the generated scenario is semantically close to what actually occurs. Let $\hat{\mathbf{s}}^{\mathrm{scn}}_{\mathrm{oracle}}$ be an oracle scenario description obtained by summarizing the ground-truth future $\mathbf{x}_{L+1:L+H}$ with the same prompt. This oracle is never used during training or inference. We compute SCS using a separate sentence-level diagnostic encoder $\operatorname{Enc}{\mathrm{diag}}(\cdot)$:
\begin{equation} 
\label{}
\mathrm{SCS}(\mathbf{s}^{\mathrm{scn}}) = \cos\!\left( \mathrm{Enc}_{\mathrm{text}}(\mathbf{s}^{\mathrm{scn}}), \mathrm{Enc}_{\mathrm{text}}(\hat{\mathbf{s}}^{\mathrm{scn}}_{\mathrm{oracle}}) \right). 
\label{eq:scs}
\end{equation} 

A higher SCS indicates stronger semantic agreement between the generated scenario and the realized future. We use SCS alongside distributional metrics (MMD, Fréchet distance) in our empirical analysis to verify that generated scenarios carry meaningful signal above noise baselines (Section~\ref{subsec:ablation}).

\paragraph{Anchor Guidance Agent}
The scenario description provides a coarse future prior, but it does not directly specify where abrupt changes may occur. We therefore introduce an \emph{Anchor Guidance Agent}, which converts the available context into sparse \emph{anchor points} for inference-time guidance.

Given $\mathbf{x}_{1:L}$, $\mathbf{s}_{1:L}^{\mathrm{ctx}}$, and $\mathbf{s}^{\mathrm{scn}}$, the Anchor Guidance Agent outputs a set of $M$ anchor points
\(
\mathcal{A} = \{(t_j, l_j, u_j, w_j)\}_{j=1}^{M},
\)
where $t_j \in \{L+1,\dots,L+H\}$ is a future timestep, $[l_j, u_j]$ is a plausible value interval, and $w_j$ is a confidence weight. We define
\begin{align}
w_j
=
\mathrm{clip}\!\left(
c\left(1 + \frac{1}{\varepsilon + |u_j - l_j|}\right),
0, w_{\max}
\right),
\end{align}
so that tighter intervals yield stronger guidance. These anchor points provide time-localized structure that complements the global scenario description and are used during inference to refine sampled trajectories.

\subsection{Multimodal Diffusion Transformer}
\paragraph{Architecture}
We choose a \texttt{Transformer}-based denoising architecture that enhances the model's ability to capture the conditioning signal for time series.
Given an input time series, we first apply \texttt{RevIN}~\cite{DBLP:conf/iclr/KimKTPCC22} to reduce temporal distribution shift and outlier effects. 

The normalized series is then transformed into tokens by a tokenization module $\Phi_{\mathrm{ts}}(\cdot)$, which incorporates the diffusion step embedding $t$: \(\mathbf{Z}_0 = \Phi_{\mathrm{ts}}\!\left(\mathrm{RevIN}([\mathbf{x}_{1:L};\mathbf{x}^{(t)}_{L+1:L+H}]),t\right).\)

\paragraph{Token Distiller}
The textual conditions can be long, especially when many historical summaries are available. To reduce computation, we compress long contextual token sequences $\mathbf{C}^{\text{hist}}$ and $\mathbf{C}^{\text{scn}}$ into a compact window of $Q$ distilled tokens, where $Q\ll L$. 
We introduce learnable prototype queries $\mathbf{R}\in\mathbb{R}^{Q\times D}$ and obtain the distilled context through cross-attention:
\begin{align}
\mathbf{C}^{\text{ctx}}_{\text{distill}}
&=\operatorname{CrossAttn}(\mathbf{R},\mathbf{C}^{\text{ctx}})
\in\mathbb{R}^{Q\times D}, \\
\mathbf{C}^{\text{scn}}_{\text{distill}}
&=\operatorname{CrossAttn}(\mathbf{R},\mathbf{C}^{\text{scn}})
\in\mathbb{R}^{Q\times D}.
\end{align}

Here, $\mathbf{R}$ acts as a set of adaptive semantic anchors that selectively aggregate information from the full context. Padding tokens are masked during attention, and fully masked sequences are mapped to zero outputs for stability. The distilled tokens are then integrated into the \texttt{Transformer} decoder layers through a cross-attention module. This module, therefore, replaces the original long context with fixed-size distilled representations, preserving salient global information while reducing the computational cost of downstream reasoning.

\paragraph{Condition Injection} 
To inject conditional information during reverse denoising, the model uses adaptive layer normalization (AdaLN)~\cite{DBLP:conf/iclr/LiZYLZ025}. Specifically, the diffusion-step embedding and the distilled contextual representations generate scale and shift parameters for \texttt{Transformer} hidden states. This allows the denoising network to adapt its computation at each diffusion step according to both temporal noise level and multimodal context.

\begin{algorithm}[t]
\caption{Asymmetric conditioning in \framework}
\label{alg:asymmetric_conditioning}
\small
\begin{algorithmic}[1]
\Require $\mathbf{x}_{1:L}$, $\mathbf{x}^{(t)}_{L+1:L+H}$,
$\mathbf{s}^{\mathrm{ctx}}_{1:L}$, $\mathbf{s}^{\mathrm{scn}}$, step $t$
\Ensure $\hat{\mathbf{x}}^{(0)}_{L+1:L+H}$

\State \textbf{Timestep tokenization:}
\State $\mathbf{Z}_0 \leftarrow
\Phi_{\mathrm{ts}}(
\operatorname{RevIN}([\mathbf{x}_{1:L};
\mathbf{x}^{(t)}_{L+1:L+H}]),t)$

\State \textbf{Text encoding:}
\State $\bar{\mathbf{C}}_h \leftarrow
\operatorname{Distill}(
\operatorname{Enc}_{\mathrm{text}}(\mathbf{s}^{\mathrm{ctx}}_{1:L}))$
\State $\bar{\mathbf{C}}_s \leftarrow
\operatorname{Distill}(
\operatorname{Enc}_{\mathrm{text}}(\mathbf{s}^{\mathrm{scn}}))$

\For{$\ell=0,\ldots,D-1$}
    \State Split tokens:
    \State $\mathbf{H}_\ell \leftarrow \mathbf{Z}_{\ell,1:L},
    \quad
    \mathbf{F}_\ell \leftarrow \mathbf{Z}_{\ell,L+1:L+H}$

    \State Inject historical context:
    \State $\tilde{\mathbf{H}}_\ell
    \leftarrow
    \mathbf{H}_\ell+\psi_h(\bar{\mathbf{C}_h})$

    \State Inject scenario guidance:
    \State $\tilde{\mathbf{F}}_\ell
    \leftarrow
    \mathbf{F}_\ell+
    \operatorname{CA}(\mathbf{F}_\ell,
    \bar{\mathbf{C}}_s,\bar{\mathbf{C}}_s)$

    \State Joint denoising update:
    \State $\mathbf{Z}_{\ell+1}
    \leftarrow
    \mathcal{B}_\ell(
    [\tilde{\mathbf{H}}_\ell;\tilde{\mathbf{F}}_\ell],t)$
\EndFor

\State \textbf{Future reconstruction:}
\State $\hat{\mathbf{x}}^{(0)}_{L+1:L+H}
\leftarrow
\operatorname{RevIN}^{-1}
(\operatorname{Head}(\mathbf{Z}_{D,L+1:L+H}))$

\State \Return $\hat{\mathbf{x}}^{(0)}_{L+1:L+H}$
\end{algorithmic}
\end{algorithm}

The detailed architecture is summarized in Alg.~\ref{alg:asymmetric_conditioning}. Here, $\Phi_{\mathrm{ts}}$ denotes time-series tokenization, $\psi_h$ is the additive history-side injection MLP, $\operatorname{CA}$ denotes cross-attention, $\mathcal{B}_{\ell}$ denotes an \texttt{AdaLN} \texttt{Transformer} block, and $\operatorname{Distill}$ compresses long contextual token sequences using learnable prototype queries.

Finally, the resulting patch tokens are processed by a stack of \texttt{Transformer} encoder layers with self-attention and feed-forward blocks, and a lightweight flatten-linear decoder maps the final hidden representations back to the reconstructed denoised time series.

\subsection{Anchor Blended Sampling}
\label{sec:blended_sampling}

After standard reverse diffusion, we obtain an initial forecast $\mathbf{x}^{(0)}_\theta$, which is used as the source trajectory for a post-hoc editing stage. 
Anchor Blended Sampling refines this source trajectory by editing anchor-relevant regions while preserving the remaining temporal structure. Unlike image-based blended diffusion, which often uses CLIP guidance~\cite{DBLP:conf/cvpr/AvrahamiLF22}, our refinement is defined by the distance-to-band objective $\mathcal{L}_{\mathrm{anch}}(\mathbf{x};\mathcal{A})$. 

Alg.~\ref{alg:energy_blended_refinement} summarizes the complete refinement procedure. Given anchors $\mathcal{A}=\{(t_j,l_j,u_j,w_j)\}_{j=1}^{M}$, we evaluate whether the trajectory satisfies each anchor interval within a local temporal window by a refinement loss:
\begin{equation}
    \small
\mathcal{L}_{\mathrm{anch}}(\mathbf{x};\mathcal{A})
=
-\tau \sum_{j=1}^{M} w_j\,
\mathrm{LogSumExp}\!\Big(
-\frac{d(\mathbf{x}_{t_j-r:t_j+r}, [l_j,u_j])}{\tau}
\Big),
\end{equation}
where $d(\cdot;[l_j,u_j])= \left(\max(0,l_j-x)+\max(0,x-u_j)\right)^2$ penalizes values outside the anchor band and $r$ allows small temporal shifts. 

\begin{algorithm}[t]
\caption{Anchor Blended Sampling}
\label{alg:energy_blended_refinement}
\small
\begin{algorithmic}[1]
\Require diffusion model $\mathbf{x}_\theta$, anchors $\mathcal{A}$, mask $\mathbf{m}$, edit steps $T_e$, scales $\{\gamma_t\}_{t=1}^{T_e}$
\Ensure refined sample $\mathbf{x}^{(0)}$
\State Sample $\mathbf{x}^{(T)}\sim\mathcal{N}(\mathbf{0},\mathbf{I})$
\For{$t=T,\ldots,1$}
    \State Draw $\mathbf{x}^{(t-1)}$ using the reverse diffusion kernel.
\EndFor
\State Set source trajectory $\mathbf{x}^{(0)}_\theta\leftarrow\mathbf{x}^{(0)}$
\State Noise the source to level $T_e$: $\mathbf{x}^{(T_e)}_\theta\sim q(\mathbf{x}^{(T_e)}\mid\mathbf{x}^{(0)}_\theta)$
\For{$t=T_e,\ldots,1$}
    \State Estimate posterior parameters $\mu_t,\Sigma_t$
    \State $g_t\leftarrow-\nabla_{\mathbf{x}^{(t)}}\mathcal{L}_{\mathrm{anch}}(\mathbf{x}_\theta(\mathbf{x}^{(t)},t);\mathcal{A})$
    \State $\tilde{\mathbf{x}}^{(t-1)}\sim\mathcal{N}(\mu_t+\gamma_t\Sigma_t g_t,\Sigma_t)$
    \State Draw $\mathbf{x}^{(t-1)}_\theta\sim q(\mathbf{x}^{(t-1)}\mid\mathbf{x}^{(0)}_\theta)$
    \State $\mathbf{x}^{(t-1)}\leftarrow\mathbf{m}\odot\tilde{\mathbf{x}}^{(t-1)}+(1-\mathbf{m})\odot\mathbf{x}^{(t-1)}_\theta$
\EndFor
\State \Return $\mathbf{x}^{(0)}$
\end{algorithmic}
\end{algorithm}

\begin{remark}[Connection to classifier guidance]
Anchor Blended Sampling is an instance of classifier-guided diffusion~\cite{DBLP:conf/nips/DhariwalN21}. Define the implicit ``anchor classifier'' $p(\mathcal{A}\mid\mathbf{x}^{(t)})\propto\exp\!\bigl(-\mathcal{L}_{\mathrm{anch}}(\mathbf{x}_\theta(\mathbf{x}^{(t)},t);\mathcal{A})\bigr)$.
The score of this classifier with respect to $\mathbf{x}^{(t)}$ is
\(
\nabla_{\mathbf{x}^{(t)}}\log p(\mathcal{A}\mid\mathbf{x}^{(t)})
= -\nabla_{\mathbf{x}^{(t)}}\mathcal{L}_{\mathrm{anch}} = g_t,
\)
so the guided posterior mean $\mu_t + \gamma_t\Sigma_t g_t$ is exactly the classifier-guided update of \citet{DBLP:conf/nips/DhariwalN21} with guidance scale $\gamma_t$.
Here, $\mathcal{L}_{\mathrm{anch}}$ is an energy function, while $-\mathcal{L}_{\mathrm{anch}}$ plays the role of the implicit log-classifier up to an additive constant.
The blending step with $\mathbf{m}$ confines this guidance to anchor-relevant regions, leaving non-anchor timesteps undisturbed---an inpainting-style constraint analogous to \citet{DBLP:conf/cvpr/AvrahamiLF22} transposed from pixels to time.
\end{remark}

Starting from a noised version $\mathbf{x}^{(T_e)}$ of the generated source $\mathbf{x}^{(0)}_\theta$, we run a short reverse editing process. 
At each timestep $t$, the posterior estimation $\tilde{\mathbf{x}}^{(t-1)}$ is updated using the negative gradient of the anchor loss:
\begin{align*}
p_\theta(\tilde{\mathbf{x}}^{(t-1)}\mid\mathbf{x}^{(t)})
&= \mathcal{N}(\mu_t+\gamma_t\Sigma_t g_t,\Sigma_t), \\
g_t
&=
-\nabla_{\mathbf{x}^{(t)}}
\mathcal{L}_{\mathrm{anch}}(\mathbf{x}_\theta(\mathbf{x}^{(t)},t);\mathcal{A}).
\end{align*}

Here, $\gamma_t$ controls the guidance strength. To preserve non-anchor regions, we blend the guided latent with a noised version of $\mathbf{x}^{(0)}_\theta$:
\begin{align*}
\mathbf{x}^{(t-1)}
&=
\mathbf{m}\odot \tilde{\mathbf{x}}^{(t-1)}
+
(1-\mathbf{m})\odot \mathbf{x}^{(t-1)}_\theta, \\
\mathbf{x}^{(t-1)}_\theta &\sim \mathcal{N}(\sqrt{\bar\alpha_{t-1}}\,\mathbf{x}^{(0)}_\theta, (1-\bar\alpha_{t-1})\,\mathbf{I}),
\end{align*}
where $\mathbf{m}$ marks the anchor-relevant temporal regions.
Thus, the base diffusion model generates a plausible trajectory, and the second energy-guided blended refinement locally corrects it according to the anchor constraints without requiring retraining.

\section{Experiments}

\subsection{Experimental Settings}
\label{subsec:exp_settings}

\paragraph{Dataset}
We evaluate on the \textbf{Time-MMD} benchmark~\cite{DBLP:conf/nips/LiuXZKKSSCW0P24}, a multi-domain multimodal dataset for time series analysis. \textbf{Time-MMD} contains paired numerical and textual data from real-world domains: \textbf{Economy}, \textbf{Energy}, \textbf{Security}, \textbf{Social Good}, and \textbf{Traffic}. The series are sampled at a weekly or monthly frequency and cover long temporal ranges from 1950 to 2024. Each numerical target is associated with aligned textual facts and reports, making the benchmark suitable for evaluating forecasting models that use both numerical histories and contextual text. Following the official \texttt{MM-TSFlib} protocol~\cite{DBLP:conf/nips/LiuXZKKSSCW0P24}, we use the standard train/validation/test split with a ratio of 7:1:2.

\paragraph{Forecasting Setup}
For monthly time series, we use a lookback length of $L=36$ and prediction horizons $H\in\{6,12,18\}$. For weekly time series, we use $L=96$ and horizons $H\in\{12,24,48\}$. For brevity, we report results averaged over all prediction horizons for each domain.

\paragraph{Baselines}

We compare \framework with four groups of baselines. The first group includes numerical-only forecasters, covering Transformer-based models such as \texttt{Informer}~\citep{DBLP:conf/aaai/ZhouZPZLXZ21}, \texttt{FEDformer}~\citep{DBLP:conf/icml/ZhouMWW0022}, and \texttt{iTransformer}~\citep{DBLP:conf/iclr/LiuHZWWML24}. The second group includes LLM-prior methods, represented by \texttt{Time-LLM}~\citep{DBLP:conf/iclr/0005WMCZSCLLPW24} and \texttt{S$^2$IP-LLM}~\citep{DBLP:conf/icml/PanJGSNS24}. The third group contains multimodal forecasting methods that use paired text, including \texttt{TaTS}~\citep{DBLP:journals/corr/abs-2502-08942} and \texttt{MM-TSF}~\citep{DBLP:conf/nips/LiuXZKKSSCW0P24}. The fourth group consists of probabilistic diffusion-based models such as \texttt{CSDI}~\citep{DBLP:conf/nips/TashiroSSE21} and \texttt{TMDM}~\citep{DBLP:conf/iclr/LiCH0SZ24}; for these models, we additionally report CRPS. We follow the \texttt{TSLib}~\citep{DBLP:conf/iclr/WuHLZ0L23} and \texttt{MM-TSFlib}~\citep{DBLP:conf/nips/LiuXZKKSSCW0P24} settings when available and tune hyperparameters on the validation set. Full baseline citations are given in Table~\ref{tab:main_results}.

\paragraph{Metrics}
We evaluate the deterministic forecasting accuracy using Mean Squared Error (MSE) and Mean Absolute Error (MAE) on standardized time series. For stochastic forecasting models, we additionally report the Continuous Ranked Probability Score (CRPS), which measures the quality of the predictive distribution. Lower values indicate better performance for all metrics.

\paragraph{Hyperparameters and Implementation Details}

All models are implemented in PyTorch and trained on a single NVIDIA H100 GPU. Across the \textbf{Time-MMD} experiments, the diffusion forecaster uses $D=6$ residual blocks, hidden dimension $d\in\{64,128,256\}$, and 8 attention heads. Models are trained with Adam for 60--150 epochs, with batch size selected from $\{16,32\}$, learning rate from $\{10^{-4},5\times10^{-4}\}$, and weight decay $10^{-6}$. Forecast trajectories are generated using \texttt{DDIM}~\citep{DBLP:conf/iclr/SongME21} with 50 sampling steps.


The Historical Context, Scenario, and Anchor Guidance Agents use \texttt{Gemini 2.5 Flash}. Agent outputs are generated offline and cached before training and inference. Text inputs are truncated to 512 tokens and encoded with a frozen \texttt{bert-base-uncased} encoder, yielding $d_{\mathrm{emb}}=768$ representations. To avoid temporal leakage, each forecast origin uses only documents and textual representations aligned with observed timesteps.


For Anchor Blended Sampling, we retain at most five anchor intervals, construct the edit mask with radius $r=2$, set the soft-min temperature to $\tau=0.10$, and perform $N_{\mathrm{edit}}=6$ guided \texttt{DDIM} reverse-editing steps. The anchor gradient is mask-restricted, clipped to unit norm, and applied with guidance scale $\gamma=50$; non-anchor regions are blended with the noised source forecast.

\subsection{Main Results}
\label{subsec:main_results}

\begin{table*}[t]
\footnotesize
\setlength{\tabcolsep}{5pt} 
\centering
\caption{Overall results on \textbf{Time-MMD}. The best, runner-up, and third-best mean results are highlighted in \firstlegend{red}, \secondlegend{blue}, and \thirdlegend{bold}, respectively. \textbf{Hor. 1st} counts the number of first-place results over all horizon-domain-metric entries, while \textbf{Ovr. 1st} counts first-place results after averaging over horizons within each domain and metric.}
\label{tab:main_results}
\begin{tabular}{l c c cccccccccc}
\toprule
\multirow{2}{*}{\textbf{Models}} & 
\multirow{2}{*}{\textbf{Hor. 1st}} & 
\multirow{2}{*}{\textbf{Ovr. 1st}}
& \multicolumn{2}{c}{\textbf{Economy}} 
& \multicolumn{2}{c}{\textbf{Energy}} 
& \multicolumn{2}{c}{\textbf{Security}} 
& \multicolumn{2}{c}{\textbf{Social Good}} 
& \multicolumn{2}{c}{\textbf{Traffic}} \\
\cmidrule(lr){4-5} 
\cmidrule(lr){6-7} 
\cmidrule(lr){8-9} 
\cmidrule(lr){10-11} 
\cmidrule(lr){12-13}
& & & 
\textbf{MSE} & \textbf{MAE} & 
\textbf{MSE} & \textbf{MAE} & 
\textbf{MSE} & \textbf{MAE} & 
\textbf{MSE} & \textbf{MAE} & 
\textbf{MSE} & \textbf{MAE} \\ 
\midrule

\texttt{Informer}~\citep{DBLP:conf/aaai/ZhouZPZLXZ21}     & 0 & 0 & 0.891 & 0.774 & 0.456 & 0.525 & 127.556 & 6.595 & 0.973 & 0.599 & 0.248 & 0.405 \\

\texttt{Reformer}~\citep{DBLP:conf/iclr/KitaevKL20}    & 0 & 0 & 1.036 & 0.853 & 0.676 & 0.632 & 122.835 & 6.285 & 1.046 & 0.647 & 0.292 & 0.443 \\
\texttt{Autoformer}~\citep{DBLP:conf/nips/WuXWL21}  & 0 & 0 & 0.340 & 0.465 & 0.478 & 0.527 & 112.782 & 5.314 & 1.618 & 0.810 & 0.240 & 0.301 \\
\texttt{FEDformer}~\citep{DBLP:conf/icml/ZhouMWW0022}   & 2 & 0 & 0.286 & 0.410 & 0.394 & 0.457 & 113.725 & 5.382 & 1.238 & 0.666 & 0.231 & 0.271 \\
\texttt{PatchTST}~\citep{DBLP:conf/iclr/NieNSK23}    & 5 & 3 & \third{0.255} & \third{0.387} & \first{0.203} & \first{0.326} & 91.734 & 5.429 & 1.100 & \third{0.559} & \second{0.102} & \first{0.178} \\
\texttt{iTransformer}~\citep{DBLP:conf/iclr/LiuHZWWML24} & 5 & 2 & 0.280 & 0.398 & 0.227 & \third{0.344} & 113.248 & 5.432 & 1.235 & 0.568 & 0.208 & 0.238 \\

\texttt{PAttn}~\citep{DBLP:conf/nips/TanMGAH24}       & 3 & 0 & \second{0.237} & \second{0.376} & 0.267 & 0.388 & 83.117 & 4.956 & 1.196 & 0.569 & 0.104 & \second{0.181} \\

\texttt{DLinear}~\citep{DBLP:conf/aaai/ZengCZ023}     & 0 & 0 & 0.579 & 0.636 & 0.391 & 0.448 & 106.504 & 4.665 & 1.524 & 0.935 & 0.284 & 0.415 \\
\texttt{FiLM}~\citep{DBLP:conf/nips/ZhouMWW0YY022}        & 0 & 0 & 0.460 & 0.556 & 0.375 & 0.469 & 108.179 & 5.158 & 1.608 & 0.949 & 0.236 & 0.326 \\
\texttt{TSMixer}~\citep{DBLP:journals/corr/abs-2303-06053}    & 1 & 0 & 1.973 & 1.155 & 0.410 & 0.482 & 94.503 & 5.735 & 1.301 & 0.796 & 0.818 & 0.714 \\
\texttt{TiDE}~\citep{DBLP:journals/tmlr/DasKLMSY23}        & 0 & 0 & 0.463 & 0.548 & 0.484 & 0.518 & 91.498 & 5.548 & 1.939 & 1.045 & 0.230 & 0.378 \\

\midrule

\texttt{Time-LLM}~\citep{DBLP:conf/iclr/0005WMCZSCLLPW24}     & 0 & 0 & 0.346 & 0.469 & 0.464 & 0.491 & 79.945 & 4.785 & 1.924 & 1.071 & 0.195 & 0.330 \\
\texttt{S$^2$IP-LLM}~\citep{DBLP:conf/icml/PanJGSNS24} & 4 & 0 & 0.273 & 0.417 & \second{0.224} & \second{0.343} & \second{76.184} & \third{4.381} & 1.025 & 0.594 & 0.191 & 0.310 \\

\midrule

\texttt{TaTS}~\citep{DBLP:journals/corr/abs-2502-08942}        & 7 & 3 & 0.924 & 0.732 & 0.457 & 0.540 & 124.457 & 6.368 & \first{0.886} & \second{0.541} & 0.184 & 0.312 \\
\texttt{MM-TSF}~\citep{DBLP:conf/nips/LiuXZKKSSCW0P24}      & 0 & 0 & 0.816 & 0.737 & 0.397 & 0.481 & 127.460 & 6.573 & \third{0.959} & 0.580 & 0.229 & 0.382 \\

\midrule

\texttt{CSDI}~\citep{DBLP:conf/nips/TashiroSSE21}        & 4 & 3 & 1.396 & 0.943 & 0.545 & 0.531 & 96.372 & 6.092 & \second{0.908} & \first{0.491} & 0.124 & 0.256 \\
\texttt{TMDM}~\citep{DBLP:conf/iclr/LiCH0SZ24}        & 3 & 1 & 1.038 & 0.765 & 0.320 & 0.398 & \third{76.270} & \first{4.177} & 1.305 & 0.673 & \third{0.103} & 0.187 \\
\texttt{NsDiff}~\citep{DBLP:conf/icml/YeXG25}      & 1 & 0 & 1.781 & 1.200 & 0.245 & 0.403 & 100.422 & 6.356 & 3.109 & 1.527 & 0.255 & 0.437 \\
\texttt{TimeDiff}~\citep{DBLP:conf/icml/ShenK23}    & 0 & 0 & 3.372 & 1.718 & 1.015 & 0.799 & 102.157 & 6.541 & 2.234 & 1.188 & 2.488 & 1.521 \\

\midrule

\framework           & \textbf{13} & \textbf{4} & \first{0.216} & \first{0.353} & \third{0.225} & 0.367 & \first{74.802} & \second{4.261} & 1.313 & 0.707 & \first{0.099} & \third{0.183} \\

\bottomrule
\end{tabular}%
\end{table*}

Table~\ref{tab:main_results} reports average MSE and MAE over prediction horizons across the five \textbf{Time-MMD} domains. Overall, \framework achieves the strongest horizon-level performance, with the largest number of MSE/MAE wins across horizons and domains. The gains are most evident in event-driven domains, especially \textbf{Economy} and \textbf{Security}, while \framework also shows competitive performance on \textbf{Traffic}. These results suggest that noisy documents become more effective for forecasting when transformed into explicit scenario-level signals, rather than fused as unstructured text. Compared with numerical-only and LLM-prior baselines, \framework benefits from using LLM agents for contextual reasoning instead of direct numerical prediction. The agents extract historical evidence, generate scenario descriptions, and produce anchor points that guide a dedicated probabilistic forecaster. This separation of historical context, scenario-level guidance, and anchor-based refinement leads to stronger performance when text contains actionable event signals.


\begin{table}[t]
\centering
\caption{Average CRPS over prediction horizons for representative diffusion-based probabilistic forecasting models. We additionally report the average over event-driven domains, where textual scenarios provide actionable future signals.}
\label{tab:crps_result}
\footnotesize
\setlength{\tabcolsep}{4pt}
\renewcommand{\arraystretch}{1.05}
\begin{tabular}{lcccccc}
\toprule
\textbf{Model} &
\textbf{Econ.} &
\textbf{Ener.} &
\textbf{Sec.} &
\textbf{Soc.} &
\textbf{Traf.} &
\textbf{Event Avg.} \\
\midrule
\texttt{CSDI}     & \third{0.231} & 0.138 & \third{0.805} & \first{0.109} & \third{0.043} & \third{0.392} \\
\texttt{TMDM}     & \second{0.190} & \first{0.099} & \first{0.511} & \second{0.156} & \first{0.030} & \second{0.267} \\
\texttt{NSDiff}   & 0.317 & \second{0.104} & 0.852 & 0.375 & 0.064 & 0.424 \\
\texttt{TimeDiff} & 0.436 & 0.197 & 0.883 & 0.266 & 0.251 & 0.506 \\
\midrule
\framework        & \first{0.094} & \third{0.112} & \second{0.584} & \third{0.189} & \second{0.036} & \first{0.263} \\
\bottomrule
\end{tabular}%
\end{table}


Table~\ref{tab:crps_result} reports CRPS for representative diffusion-based probabilistic forecasting models. \framework achieves the best event-domain average CRPS, indicating that hierarchical scenario guidance can also improve predictive distributions when textual evidence is informative. The improvement is not uniform across all domains, as strong numerical or diffusion-based baselines remain competitive in some settings. This domain-dependent behavior supports our main motivation: scenario-level guidance is most beneficial when external textual evidence provides actionable signals about future dynamics.


\subsection{Ablation Studies}
\label{subsec:ablation}

\paragraph{Contribution of Components}

\begin{figure*}[t]
\scriptsize
\centering

\begin{subfigure}{0.24\linewidth}
    \begin{tikzpicture}
        \begin{axis}[
            ybar,
            ymin=0,
            ymax=0.45,
            ytick={0,0.1,0.2,0.3,0.4},
            bar width=4pt,
            width=1.15\linewidth,
            height=0.85\linewidth,
            ylabel={Error},
            symbolic x coords={Ours,-future,-intrinsic,-anchor},
            xticklabels={\texttt{Full},\texttt{-Scn.},\texttt{-Hist.},\texttt{-Anchor}},
            xtick=data,
            xticklabel style={rotate=35, anchor=east},
            legend style={at={(2.5,1.15)}, anchor=south, legend columns=2},
        ]
            \addplot[ybar, fill=blue!55] table[
                col sep=comma,
                x=variant,
                y=MSE_economy,
            ] {data/main_ablation_raw.csv};
            \addlegendentry{MSE}

            \addplot[ybar, fill=orange!70] table[
                col sep=comma,
                x=variant,
                y=MAE_economy,
            ] {data/main_ablation_raw.csv};
            \addlegendentry{MAE}
        \end{axis}
    \end{tikzpicture}
    \caption{\textbf{Economy}}
    \label{subfig:ablation-economy}
\end{subfigure}
\hfill
\begin{subfigure}{0.24\linewidth}
    \begin{tikzpicture}
        \begin{axis}[
            ybar,
            ymin=0,
            ymax=0.5,
            ytick={0,0.1,0.2,0.3,0.4,0.5},
            bar width=4pt,
            width=1.15\linewidth,
            height=0.85\linewidth,
            ylabel={Error},
            symbolic x coords={Ours,-future,-intrinsic,-anchor},
            xticklabels={\texttt{Full},\texttt{-Scn.},\texttt{-Hist.},\texttt{-Anchor}},
            xtick=data,
            xticklabel style={rotate=35, anchor=east},
        ]
            \addplot[ybar, fill=blue!55] table[
                col sep=comma,
                x=variant,
                y=MSE_energy,
            ] {data/main_ablation_raw.csv};

            \addplot[ybar, fill=orange!70] table[
                col sep=comma,
                x=variant,
                y=MAE_energy,
            ] {data/main_ablation_raw.csv};
        \end{axis}
    \end{tikzpicture}
    \caption{\textbf{Energy}}
    \label{subfig:ablation-energy}
\end{subfigure}
\hfill
\begin{subfigure}{0.24\linewidth}
    \begin{tikzpicture}
        \begin{axis}[
            ybar,
            ymin=0,
            ymax=100,
            ytick={0,20,40,60,80,100},
            bar width=4pt,
            width=1.15\linewidth,
            height=0.85\linewidth,
            ylabel={Error},
            symbolic x coords={Ours,-future,-intrinsic,-anchor},
            xticklabels={\texttt{Full},\texttt{-Scn.},\texttt{-Hist.},\texttt{-Anchor}},
            xtick=data,
            xticklabel style={rotate=35, anchor=east},
        ]
            \addplot[ybar, fill=blue!55] table[
                col sep=comma,
                x=variant,
                y=MSE_security,
            ] {data/main_ablation_raw.csv};

            \addplot[ybar, fill=orange!70] table[
                col sep=comma,
                x=variant,
                y=MAE_security,
            ] {data/main_ablation_raw.csv};
        \end{axis}
    \end{tikzpicture}
    \caption{\textbf{Security}}
    \label{subfig:ablation-security}
\end{subfigure}
\hfill
\begin{subfigure}{0.24\linewidth}
    \begin{tikzpicture}
        \begin{axis}[
            ybar,
            ymin=0,
            ymax=2.5,
            ytick={0,0.5,1.0,1.5,2.0,2.5},
            bar width=4pt,
            width=1.15\linewidth,
            height=0.85\linewidth,
            ylabel={Error},
            symbolic x coords={Ours,-future,-intrinsic,-anchor},
            xticklabels={\texttt{Full},\texttt{-Scn.},\texttt{-Hist.},\texttt{-Anchor}},
            xtick=data,
            xticklabel style={rotate=35, anchor=east},
        ]
            \addplot[ybar, fill=blue!55] table[
                col sep=comma,
                x=variant,
                y=MSE_socialgood,
            ] {data/main_ablation_raw.csv};

            \addplot[ybar, fill=orange!70] table[
                col sep=comma,
                x=variant,
                y=MAE_socialgood,
            ] {data/main_ablation_raw.csv};
        \end{axis}
    \end{tikzpicture}
    \caption{\textbf{Social Good}}
    \label{subfig:ablation-socialgood}
\end{subfigure}
\vspace{-4pt}
\caption{Ablation study across four domains. Each subfigure reports the absolute MSE and MAE of the full \framework model and variants without the Scenario Agent, Historical Context Agent, and Anchor Guidance Agent.}
\label{fig:ablation_domains}
\vspace{-17pt}
\end{figure*}
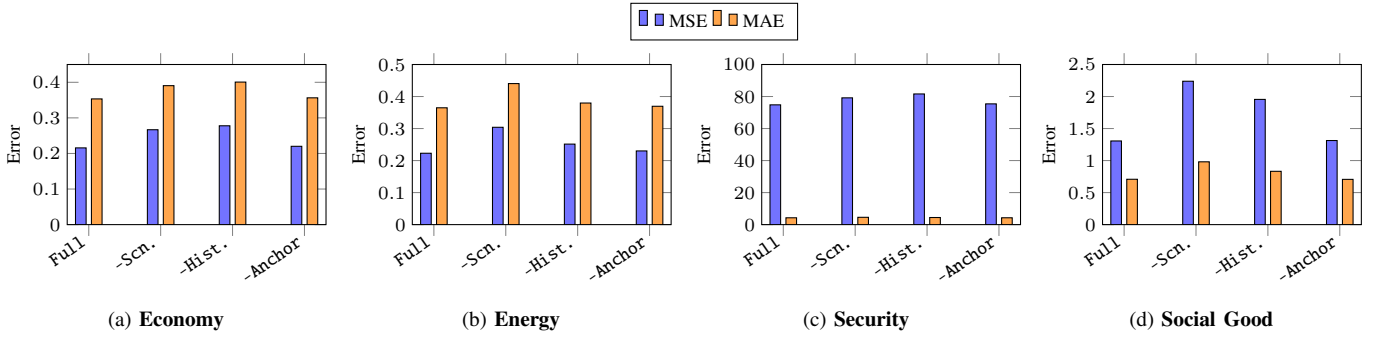

We evaluate the contribution of the three agent modules in \framework by removing one component at a time. Fig.~\ref{fig:ablation_domains} compares the full model with three variants: w/o Historical Context Agent (-His.), w/o Scenario Agent (-Scn.), and w/o Anchor Guidance Agent (-Anchor). Each bar reports the absolute MSE/MAE of the corresponding variant, so larger values indicate worse forecasting accuracy.

The results show that all three modules contribute to the final performance. Removing the Historical Context Agent causes the largest degradation in most domains, indicating that document-aligned historical evidence is important for grounding the forecast. Removing the Scenario Agent also hurts performance, especially in event-driven domains, confirming the value of forecast-horizon scenario descriptions. Removing the Anchor Guidance Agent leads to a smaller but consistent drop, suggesting that anchor guidance mainly acts as an inference-time refinement mechanism. Overall, the ablation results support the hierarchical design of \framework: historical summaries provide grounding, scenario descriptions provide future guidance, and anchor points provide localized refinement.

\paragraph{Event-driven Subset Analysis}

\begin{table}[t]
\centering
\caption{Average MSE/MAE rank across event-driven and non-event-driven domains. Ranks are computed over all models in Table~\ref{tab:main_results} for each reported MSE and MAE column, while this table reports representative strong baselines for compactness. Lower rank is better.}
\label{tab:event_driven}
\footnotesize
\setlength{\tabcolsep}{4.5pt}
\renewcommand{\arraystretch}{1.05}
\begin{tabular}{lcc}
\toprule
\textbf{Model} &
\textbf{Event-driven} &
\textbf{Non-event-driven} \\
\midrule
\texttt{TaTS}             & 15.50 & \third{4.75} \\
\texttt{S$^2$IP-LLM}      & \second{3.17} & 7.00 \\
\texttt{MM-TSF}           & 14.17 & 8.25 \\
\midrule
\texttt{PatchTST}         & \third{4.17} & \first{3.25} \\
\texttt{iTransformer}     & 7.00 & 6.75 \\
\texttt{PAttn}            & 4.33 & \third{4.75} \\
\midrule
\texttt{CSDI}             & 15.33 & \second{3.50} \\
\texttt{TMDM}             & 7.83 & 7.50 \\
\midrule
\framework                & \first{2.00} & 7.25 \\
\bottomrule
\end{tabular}
\end{table}

To assess where scenario-level guidance contributes most, we partition the \textbf{Time-MMD} domains into event-driven domains (\textbf{Economy}, \textbf{Energy}, and \textbf{Security}) and non-event-driven domains (\textbf{Social Good} and \textbf{Traffic}). Table~\ref{tab:event_driven} reports the average rank over both MSE and MAE within each partition.
\framework achieves the strongest mean rank on event-driven domains, outperforming representative numerical-only baselines such as \texttt{PatchTST} and \texttt{PAttn}, the LLM-prior method \texttt{S$^2$IP-LLM}, and multimodal or diffusion-based baselines such as \texttt{TaTS} and \texttt{TMDM}. This result indicates that scenario-level guidance is most beneficial when external events and textual narratives provide actionable clues about future dynamics.

In non-event-driven domains, the advantage of \framework becomes less pronounced. Strong numerical-only, multimodal, and diffusion-based baselines achieve better average ranks, suggesting that these domains are more dominated by regular temporal patterns and that textual evidence provides fewer actionable signals. This contrast supports our central motivation: scenario-level guidance is most useful when future dynamics are shaped by external events that are not fully recoverable from historical values alone.


\paragraph{Quality of LLM-generated Context}

\begin{table}[t]
\centering
\footnotesize
\caption{Quality of LLM-generated context in embedding space on \textbf{Energy}. \textit{Coarse} uses history only; \textit{Oracle} uses future (reference only). \textit{Shuffle/Random} breaks alignment by shuffling contexts or sampling unrelated text.}
\label{tab:llm_quality}
\setlength{\tabcolsep}{7pt}
\begin{tabular}{lp{1.2cm}p{2cm}}
\toprule
\textbf{Metric} & \textbf{Oracle vs. Coarse} & \textbf{Oracle vs. Shuffle/Random} \\
\midrule
Cosine similarity $\uparrow$  & 0.632 & 0.587 (Shuffle) \\
MMD $\downarrow$              & 0.376 & 0.579 (Random) \\
Fr\'echet distance $\downarrow$ & 0.432 & 1.532 (Random) \\
\bottomrule
\end{tabular}
\end{table}

To verify whether the generated context carries meaningful signal beyond noise, we compare three text sources in SCS (see Eq.~\ref{eq:scs}) on \textbf{Energy}: (i) \textit{Coarse}, the history-only scenario generated by the Scenario Agent; (ii) \textit{Oracle}, a reference scenario derived from ground-truth future values and used only for analysis; and (iii) misaligned baselines, \textit{Shuffle/Random}, which break alignment by shuffling contexts across samples or replacing them with unrelated text. Table~\ref{tab:llm_quality} reports semantic and distributional agreement in the diagnostic \texttt{SentenceTransformer}~\citep{DBLP:conf/emnlp/ReimersG19} embedding space. The generated coarse scenarios are consistently closer to the oracle reference than the misaligned baselines across all metrics, indicating that the Scenario Agent produces contextual signals distinguishable from random or misaligned text.


\paragraph{Anchor Quality Analysis}

\begin{table}[t]
\centering
\footnotesize
\caption{Local anchor diagnostic for sparse future anchors. MSE@r2 and MAE@r2 measure local alignment to each anchor band within radius $r=2$. Gain MSE/MAE reports the endpoint improvement over a mean-value baseline that predicts the future as a mean trajectory.}
\label{tab:anchor_vs_model}
\begin{tabular}{lp{1.2cm}p{1.2cm}p{1.5cm}p{1.55cm}}
\toprule
\textbf{Dataset} 
& \textbf{Anchor NMSE@r2} 
& \textbf{Anchor NMAE@r2}
& \textbf{Gain NMSE (\%)}
& \textbf{Gain NMAE (\%)} \\
\midrule
\textbf{Economy}    & 0.121 & 0.229 & 47.63 & 37.50 \\
\textbf{Energy}     & 0.393 & 0.427 & -89.70 & -24.19 \\
\textbf{Social Good} & 0.498 & 0.408 & 60.48 & 49.99 \\
\textbf{Traffic}    & 0.040 & 0.100 & 59.58 & 39.96 \\
\bottomrule
\end{tabular}
\end{table}


For each sparse anchor, we measure whether the predicted trajectory reaches the anchor band within a local radius $r=2$. MSE@r2 and MAE@r2 measure local anchor alignment, while the gain metrics compare endpoint error against a flat mean-value baseline.


Table~\ref{tab:anchor_vs_model} shows that anchors can provide useful local signals. \textbf{Economy} and \textbf{Traffic} have lower anchor errors and positive endpoint gains, suggesting that their anchors capture meaningful future deviations. In contrast, \textbf{Energy} has higher anchor error and negative gain, indicating that inaccurate values or temporal misalignment can misguide anchor-based refinement.


This diagnostic complements the full-horizon ablation: while the ablation evaluates the forecasting benefit of anchor guidance, this analysis examines whether the extracted anchors themselves are reliable. Overall, anchors are useful as sparse local constraints for abrupt future changes, but their effectiveness depends on anchor quality and temporal alignment.

\paragraph{Parameter Sensitivity of Anchor Guidance}
\label{sec:param_sensitivity}


\begin{figure}[t]
\begin{subfigure}[b]{0.49\linewidth}
    \scriptsize
    \centering
    \begin{tikzpicture}
        \begin{axis}[
            width=1.05\linewidth,
            height=0.55*\axisdefaultheight,
            xlabel={Blending window $w$},
            ylabel={Error reduction (\%)},
            ymin=0, ymax=2.0,
            xtick={1,2,3,4,5},
            xticklabels={1,3,7,20,50},
            ytick={0,0.5,1.0,1.5,2.0},
            legend style={
                at={(0.05,1.25)},
                anchor=north west,
                legend columns=2,
                font=\scriptsize
            },
            tick label style={font=\scriptsize},
            label style={font=\scriptsize},
            every axis plot/.append style={thick, mark=*}
        ]

        \addplot table[
            col sep=comma,
            x=id,
            y=MSE_gain_pct
        ] {data/anchor_hyper_window_sweep.csv};
        \addlegendentry{MSE}

        \addplot table[
            col sep=comma,
            x=id,
            y=MAE_gain_pct
        ] {data/anchor_hyper_window_sweep.csv};
        \addlegendentry{MAE}
        \end{axis}
    \end{tikzpicture}
    \vspace{22pt}
    \caption{Sensitivity of anchor guidance to the blending window size.}
    \label{fig:anchor_window_sensitivity}
\end{subfigure}
\begin{subfigure}[b]{0.49\linewidth}
    \scriptsize
    \centering
    \begin{tikzpicture}
        \begin{axis}[
            ybar,
            width=1.05\linewidth,
            height=0.55*\axisdefaultheight,
            ymin=-0.5,
            ymax=2.0,
            ytick={-0.5,0,0.5,1.0,1.5,2.0},
            ylabel={Error reduction (\%)},
            symbolic x coords={
                Boundary window w=1,
                Moderate windows w=3--7,
                Wide windows w=20--50,
                Dense anchors I,
                Dense anchors II
            },
            xticklabels={
                Boundary $w=1$,
                Moderate $w=3$--$7$,
                Wide $w=20$--$50$,
                Dense I,
                Dense II
            },
            xtick=data,
            xticklabel style={align=center, font=\scriptsize, rotate=35, anchor=east},
            tick label style={font=\scriptsize},
            label style={font=\scriptsize},
            bar width=6pt,
            enlarge x limits=0.15,
            legend style={
                at={(0.55,1.05)},
                anchor=south,
                legend columns=2,
                font=\scriptsize
            }
        ]

        \addplot[ybar, fill=blue!55] table[
            col sep=comma,
            x=regime,
            y=MSE_gain_pct
        ] {data/anchor_hyper_grouped_regimes.csv};
        \addlegendentry{MSE}

        \addplot[ybar, fill=orange!70] table[
            col sep=comma,
            x=regime,
            y=MAE_gain_pct
        ] {data/anchor_hyper_grouped_regimes.csv};
        \addlegendentry{MAE}

        \end{axis}
    \end{tikzpicture}
    \caption{Grouped parameter sensitivity of anchor guidance.}
    \label{fig:anchor_grouped_regimes}
\end{subfigure}
\caption{Parameter sensitivity of anchor guidance on \textbf{Economy}.}
\label{fig:anchor_guidance_sensitivity}
\vspace{-10pt}
\end{figure}
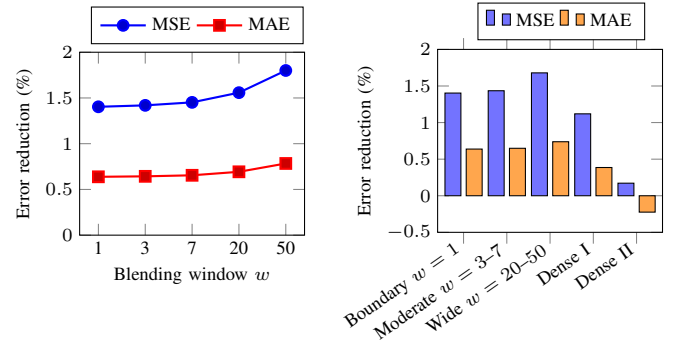



We analyze the sensitivity of the anchor guidance module on \textbf{Economy} using parameter sweeps and report the mean MSE/MAE improvement over the no-anchor baseline. As shown in Fig.~\ref{fig:anchor_guidance_sensitivity}(a), anchor guidance is generally beneficial under stable settings. Moderate and wide blending windows provide more reliable improvements than very narrow windows, suggesting that anchor information should be incorporated smoothly into the diffusion trajectory rather than imposed as overly local constraints. The grouped results in Fig.~\ref{fig:anchor_guidance_sensitivity}(b) further show that increasing the number of anchors or applying stronger anchor expansion does not consistently improve performance. This suggests that aggressive anchor augmentation may introduce noisy or overly restrictive future constraints. Overall, the sensitivity analysis indicates that anchor guidance works best with a compact anchor set and a sufficiently wide blending window.

\paragraph{Source-text perturbation}


To evaluate the sensitivity of \framework to corrupted textual signals, we perturb the source text at different ratios and report the mean forecasting errors over all prediction horizons. As shown in Fig.~\ref{fig:source_text_perturb}, the clean setting achieves the lowest MSE and MAE, indicating that textual context provides useful conditioning information. As the perturbation ratio increases, forecasting errors generally become larger, although the trend is not strictly monotonic. This suggests that \framework benefits from informative source text while retaining partial robustness to moderate textual noise.

\begin{figure}[t]
  \begin{minipage}[b]{0.49\linewidth}
    \centering
    \scriptsize
    \begin{tikzpicture}
        \begin{axis}[
            width=1.1\linewidth,
            height=0.55*\axisdefaultheight,
            xlabel={Perturbed source text (\%)},
            ylabel={Error},
            xmin=0, xmax=50,
            ymin=0.20, ymax=0.42,
            xtick={0,10,20,30,40,50},
            ytick={0.20,0.25,0.30,0.35,0.40},
            legend style={
                at={(0.95,1.05)},
                anchor=south east,
                legend columns=2,
                font=\scriptsize
            },
            tick label style={font=\scriptsize},
            label style={font=\scriptsize},
            every axis plot/.append style={thick, mark=*}
        ]

        \addplot table[
            col sep=comma,
            x=perturb_pct,
            y=MSE
        ] {data/source_text_perturb_mean.csv};
        \addlegendentry{MSE}

        \addplot table[
            col sep=comma,
            x=perturb_pct,
            y=MAE
        ] {data/source_text_perturb_mean.csv};
        \addlegendentry{MAE}

        \end{axis}
    \end{tikzpicture}
    \caption{Source-text perturbation sensitivity on \textbf{Economy}.}
    \label{fig:source_text_perturb}
  \end{minipage}
  \begin{minipage}[b]{0.48\linewidth}
    \scriptsize
    \begin{tikzpicture}
        \begin{axis}[
            width=1.1\linewidth,
            height=0.55*\axisdefaultheight,
            xlabel={Noisy anchor contamination (\%)},
            ylabel={Change vs. clean anchor (\%)},
            ymin=-5, ymax=45,
            xtick={1,2,3,4,5},
            xticklabels={0,10,30,50,100},
            ytick={0,10,20,30,40},
            legend style={
                at={(0.95,1.05)},
                legend columns=2,
                anchor=south east,
                font=\scriptsize
            },
            tick label style={font=\scriptsize},
            label style={font=\scriptsize},
            every axis plot/.append style={thick, mark=*}
        ]

        \addplot table[
            col sep=comma,
            x=id,
            y=MSE_change_pct
        ] {data/noisy_anchor_mean.csv};
        \addlegendentry{MSE}

        \addplot table[
            col sep=comma,
            x=id,
            y=MAE_change_pct
        ] {data/noisy_anchor_mean.csv};
        \addlegendentry{MAE}

        \end{axis}
    \end{tikzpicture}
    \caption{Robustness to noisy anchor contamination on \textbf{Economy}.}
    \label{fig:noisy_anchor}
  \end{minipage}
\end{figure}

\paragraph{Anchor Noise Sensitivity}




To evaluate robustness against erroneous guidance, we contaminate the anchor set with high-confidence noisy anchors sampled from the tails of each window's historical distribution. In Fig.~\ref{fig:noisy_anchor}, moderate contamination has limited impact on forecasting accuracy, indicating that Anchor Blended Sampling can tolerate a small amount of noisy guidance. In contrast, severe contamination substantially degrades performance, especially when many noisy anchors are assigned high confidence. These results suggest that anchor guidance is robust to limited noise but remains sensitive to dense, high-confidence erroneous anchors.

\subsection{Computational Cost}
We separate offline LLM preprocessing from online forecasting. The three LLM agents are called once per input instance to generate and cache the stepwise context summaries, scenario description, and anchor points, with average latencies of 8--10 seconds, 13--15 seconds, and around 15 seconds, respectively. During online forecasting, \framework loads the cached agent outputs, encodes them with a frozen text encoder, and performs diffusion sampling. On a single NVIDIA H100 GPU with batch size 64, inference takes 0.9--1.3 seconds per batch using \texttt{DDIM} with 50 denoising steps. This cost is acceptable for weekly or monthly forecasting tasks, but \framework is not designed for strict real-time or high-frequency settings, where repeated LLM preprocessing or multi-trajectory diffusion sampling may become costly.

\subsection{Case study}

\begin{figure}[t]
\centering
\begin{minipage}[t]{0.5\textwidth}
\centering
\includegraphics[width=\linewidth]{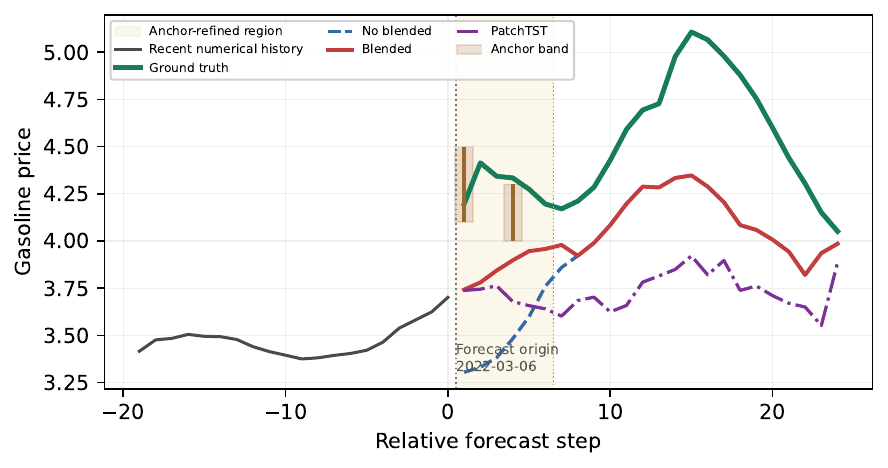}
\vspace{-1.0mm}
{\footnotesize\textbf{(a)} Mean forecasts, ground truth, and local anchor bands.}
\end{minipage}
\par\vspace{1.5mm}
\begin{minipage}[b]{0.5\textwidth}
\centering
\includegraphics[width=\linewidth]{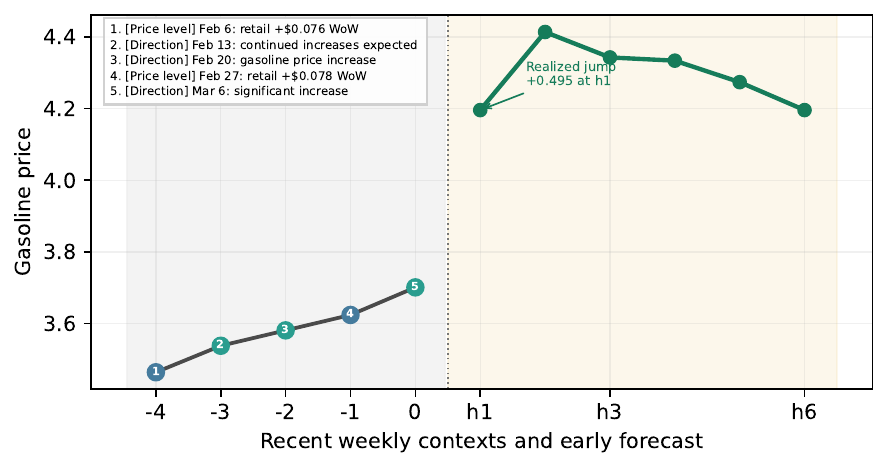}
{\footnotesize\textbf{(b)} Directionally aligned intrinsic evidence and the realized jump.}
\end{minipage}
\caption{Event-driven case in \textbf{Energy}. Historical evidence, scenario guidance, and anchor refinement jointly guide the forecast. Compared with \texttt{PatchTST}, blending improves the near-term trajectory and better follows the rising temporal pattern, although the shock magnitude remains underestimated.}
\label{fig:s00211-main-case}
\end{figure}



We present an event-driven case study from the \textbf{Energy} domain to illustrate how \framework uses contextual evidence. Fig.~\ref{fig:s00211-main-case}(a) compares the ground truth, the base diffusion forecast without blending, the anchor-refined forecast, and the \texttt{PatchTST} baseline. While \texttt{PatchTST} stays close to the historical range and misses the upward transition, Anchor Blended Sampling refines the diffusion trajectory toward the local anchor bands and better follows the realized increase.

Fig.~\ref{fig:s00211-main-case}(b) shows the recent weekly observations and stepwise context summaries before the forecast origin. The summaries consistently indicate continued gasoline-price increases, providing directional evidence for an upward regime. The realized jump shortly after the forecast origin supports their role as historical grounding evidence.

Overall, the case study illustrates the three-level reasoning process of \framework: the Historical Context Agent extracts evidence of recent price increases, the Scenario Agent forms an upward forecast-horizon scenario, and the Anchor Guidance Agent provides local bands for event-relevant future steps. Although the full shock magnitude is still underestimated, the anchor-refined trajectory better captures the direction and timing of the event-driven change.


\section{Conclusion}
\label{sec:conclusion}

We presented \framework, a hierarchical contextual reasoning framework for multimodal time series forecasting. By organizing contextual information into three levels---historical evidence (stepwise context summaries), future hypothesis (scenario description), and local constraints (anchor points)---\framework conditions a Multimodal Diffusion Transformer on structured, interpretable signals rather than raw or implicitly fused text. Anchor Blended Sampling provides inference-time trajectory refinement grounded in classifier-guided diffusion. Experiments on \textbf{Time-MMD} confirm that hierarchical scenario guidance yields the greatest benefit in event-driven domains, where textual evidence carries actionable signals about future dynamics. Future work will focus on reducing dependence on external LLMs, improving scenario quality through structured generation, and evaluating the framework across additional multimodal forecasting benchmarks.


\bibliographystyle{IEEEtran}
\bibliography{references}

\end{document}